\documentclass[conference]{IEEEtran}
\IEEEoverridecommandlockouts
\usepackage{cite}
\usepackage{amsmath,amssymb,amsfonts}
\usepackage{algorithmic}
\usepackage{graphicx,epstopdf}
\usepackage{xcolor}
\usepackage{times}
\usepackage{epsfig}
\usepackage{graphicx}
\usepackage{amsmath}
\usepackage{amssymb}
\usepackage[colorlinks,CJKbookmarks=True]{hyperref}
\usepackage{soul}
\usepackage{url}
\usepackage[utf8]{inputenc}
\usepackage[small]{caption}
\usepackage{booktabs}
\usepackage{multirow}
\usepackage{lettrine}
\usepackage{subcaption}
\usepackage{url}
\usepackage{verbatim}
\usepackage{array}
\usepackage{stix}
\usepackage{xspace}
\usepackage{verbatim}
\usepackage{color}

\def\BibTeX{{\rm B\kern-.05em{\sc i\kern-.025em b}\kern-.08em
    T\kern-.1667em\lower.7ex\hbox{E}\kern-.125emX}}

\def\modelname{Deep Comparison Network\xspace}
\def\modelnameshort{DCN}
\def\modelnamefull{\modelname (\modelnameshort)\xspace}
\def\tierIN{\textit{tiered}ImageNet}
\def\miniIN{\textit{mini}ImageNet}
\def\imagenet{ImageNet}

\newcommand{\cut}[1]{}

\newcommand{\keypoint}[1]{\vspace{0.05cm}\noindent\textbf{#1}\quad}

\begin{document}
\title{RelationNet2: Deep Comparison Columns \\for Few-Shot Learning
}

\author{
\IEEEauthorblockN{Xueting Zhang}
\IEEEauthorblockA{\textit{School of Informatics} \\
\textit{University of Edinburgh}\\
Edinburgh, UK \\
Xueting.Zhang@ed.ac.uk}
\and

\IEEEauthorblockN{Yuting Qiang}
\IEEEauthorblockA{\textit{National Key Laboratory for Software Technology} \\
\textit{Nanjing University}\\
Nanjing, China\\
qiangyuting.new@gmail.com}
\and

\IEEEauthorblockN{Flood Sung}
\IEEEauthorblockA{ 
\textit{Independent Researcher}\\
Beijing, China \\
floodsung@gmail.com}
\and

\hspace{3cm}\IEEEauthorblockN{Yongxin Yang}
\IEEEauthorblockA{\hspace{3cm}\textit{School of Informatics} \\
\hspace{3cm}\textit{University of Edinburgh}\\
\hspace{3cm}Edinburgh, UK \\
\hspace{3cm}yongxin.yang@ed.ac.uk}
\and

\IEEEauthorblockN{Timothy Hospedales}
\IEEEauthorblockA{\textit{Samsung AI Research Centre}~~~\textit{School of Informatics} \\
\hfill\textit{University of Edinburgh}\\
Cambridge, UK\hspace{2cm} Edinburgh, UK \\
t.hospedales@samsung.com\hfill t.hospedales@ed.ac.uk}
}
\maketitle

\begin{abstract}
Few-shot deep learning is a topical challenge area for scaling visual recognition to open ended growth of unseen new classes with limited labeled examples. A promising approach is based on metric learning, which trains a deep embedding to support image similarity matching. Our insight is that effective general purpose matching requires non-linear comparison of features at multiple abstraction levels. We thus propose a new deep comparison network comprised of embedding and relation modules that learn multiple non-linear distance metrics based on different levels of features simultaneously. Furthermore, to reduce over-fitting and enable the use of deeper embeddings, we represent images as distributions rather than vectors via learning parameterized Gaussian noise regularization. The resulting network achieves excellent performance on both miniImageNet and tieredImageNet.
\end{abstract}


\section{Introduction}
The ability to learn from one or few examples is an important property of human learning to function effectively in the real world. In contrast, our most successful deep learning-based approaches to recognition \cite{krizhevsky2012imagenet, he2016deep,hu2018senet} treat each learning problem as tabula-rasa, limiting their application to open-ended learning with rare data and expensive annotation (e.g., endangered species and medical images).

These observations have motivated a resurgence of interest in FSL (few-shot learning) for visual recognition \cite{vinyals2016matching,finn2017model,snell2017prototypical,qiao2017few} and beyond.  
Contemporary deep networks overfit in the few-shot regime -- even when exploiting fine-tuning \cite{yosinski2014howTransferable}, data augmentation \cite{krizhevsky2012imagenet}, or regularization \cite{srivastava2014dropout} techniques. In contrast, `Meta-learning' techniques extract transferable task agnostic knowledge from historical tasks and benefit sparse data learning of specific new target tasks. These take several forms: Fast adaptation methods enable sparse-data adaptation without overfitting -- via good initial conditions \cite{finn2017model} or learned optimizers \cite{ravi2017optimization}. Weight synthesis approaches learn a meta-network that synthesizes recognition weights given a training set  \cite{bertinetto2016feedForwardOneShot,mishra2018simple}. Deep metric learning approaches support representation \cite{koch2015siamese} and comparison \cite{vinyals2016matching,snell2017prototypical} of instances, allowing new categories to be recognized with nearest-neighbour comparison. However, existing approaches have several drawbacks including inference complexity \cite{lake2015ppi,lee2019meta}, architectural complexity \cite{munkhdalai2017meta}, the need to fine-tune on the target problem \cite{finn2017model}, or reliance on a simple linear comparison  \cite{vinyals2016matching,snell2017prototypical,lee2019meta}. 

We build on deep metric learning methods due to their architectural simplicity and instantaneous training of new categories. These methods use auxiliary training tasks to learn a deep image-embedding such that the embedded data becomes linearly separable \cite{koch2015siamese,vinyals2016matching,snell2017prototypical}. Thus the decision is non-linear in image-space, but linear in the embedding space. For learning the target task, images are simply memorized during few-shot training. But for testing the target task, query images are matched to training examples by deep embedding and similarity comparison function. Within this paradigm, the recent Relation Network \cite{yang2018learning} achieved excellent performance by learning a non-linear comparison function. Learning the embedding and non-linear comparison module jointly alleviates the reliance on the embedding's ability to generate linearly separable features.

We extend this idea of jointly learning an embedding and a non-linear distance metric with the following further insights. 
First, we introduce the notion of multiple meta-learners operating at multiple abstraction levels\cut{, where multiple layers of features have been used in other problems but just fused into a single classifier \cite{hariharan2015hypercolumn} or one shallow classifier per layer \cite{lee2015deepSupNet}}.
Concretely we train non-linear distance metrics corresponding to each embedding module in a feature hierarchy - thus covering features from simple textures to complex parts \cite{zeiler2014understandingCNN}.
Secondly, prior studies only use a single linear \cite{snell2017prototypical} or non-linear comparison \cite{yang2018learning}.
To provide the inductive bias that each layer of representation should be potentially discriminative for matching, and enable better gradient propagation \cite{huang2017densely} to each relation module, we deeply supervise \cite{lee2015deepSupNet} all the relation modules. Finally, to enable deeper embedding architectures to be used without overfitting, we design each embedding module to output a feature \emph{distribution}, thus representing each image as a distribution rather than a vector. This can be seen as an end-to-end learnable noise regularizer that performs data augmentation in semantic feature space rather than image space.

Overall RelationNet2 implements a \modelnamefull{} that can be seen as jointly learning embedding and comparison as task agnostic meta knowledge \cite{vinyals2016matching,snell2017prototypical,yang2018learning,hospedales2020metaSurvey}. It makes full use of deep networks by making comparisons with the full feature hierarchy extracted by the embedding network, and learning Gaussian noise to improve generalization. The resulting framework maintains the architecture simplicity and efficiency of other methods in this line, while providing excellent performance on both \miniIN{} and the more challenging \tierIN{} few shot learning benchmarks.

\section{Related Work}
Contemporary approaches to deep-network few-shot learning have exploited the learning-to-learn paradigm \cite{hospedales2020metaSurvey}. Auxiliary tasks are used to meta-learn some task agnostic knowledge, before exploiting this to learn the target few-sample more effectively problem. The learning-to-learn idea has a long history \cite{thrun1996lll,fei2006one,lake2015ppi}, but contemporary approaches typically cluster into three categories: Fast adaptation, weight synthesis, and metric-learning approaches. 

\keypoint{Fast Adaptation} These approaches aim to meta-learn an optimisation process that enables base models to be fine-tuned quickly and robustly. So that a base model can be updated for sparse data target problems without extensive overfitting. Effective ideas include the simply meta-learning an effective initial condition \cite{finn2017model,DBLP:journals/corr/abs-1803-02999}, and learning a recurrent neural network optimizer to replace the standard SGD learning approach \cite{ravi2017optimization}. Recent extensions also include learning per-parameter learning rates \cite{li2017meta}, and accelerating fine-tuning through solving some layers in closed form \cite{bertinetto2018closedFormMeta}. Nevertheless, these methods suffer from needing to be fine-tuned for the target problem, often generating costly higher-order gradients during meta-learning process \cite{finn2017model}, and failing to scale to deeper network architectures as shown in \cite{mishra2018simple}. They also suffer from a fixed parametric architecture. For example, once you train MAML \cite{finn2017model} for 5-way auxiliary classification problems, it is restricted to the same for target problems without being straightforwardly generalizable to a different cardinality of classification. 

\keypoint{Classifier Synthesis}
Another line of work focuses on synthesising a classifier based on the provided few-shot training data \cite{gidaris2018dynamic}. An early method in this line learned a transferrable `LearnNet' that generated convolutional weights for the base recognition network given a one-shot training example \cite{bertinetto2016feedForwardOneShot}. However, this was limited to binary classification. Conditional Neural Processes \cite{garnelo2018conditional} exploited a similar idea, but in a Bayesian framework. SNAIL obtained excellent results by embedding the training set with temporal convolutions and attention \cite{mishra2018simple}. The PPA model predicts  classification parameters given neuron activations \cite{qiao2017few}. In this case the global parameter prediction network is the task agnostic knowledge that is transferred from auxiliary categories. Compared to the fast adaptation approaches, these methods generally synthesize their classifier in a single pass, making them faster to train on the target problem. {However learning to synthesize a full classifier does entail some complexity. This process can overfit and  generalize poorly to novel target problem.}

\keypoint{Deep Metric Learning}
These approaches aim to learn a deep embedding that extracts robust features, allowing them to be classified directly with nearest neighbour type strategies in the embedding space. The deep embedding forms the task agnostic knowledge transferred from auxiliary to target tasks. Early work simply used Siamese networks \cite{koch2015siamese} to embed images, such that images of the same class are placed near each other. Matching networks \cite{vinyals2016matching} defined a differentiable nearest-neighbour loss based on cosine similarity between the support set and query embedding.  
Prototypical Networks \cite{snell2017prototypical} provide a simpler but more effective variant of this idea where the support set instances for one class are embedded as a single prototype. Their analysis showed that this leads to a linear classifier in the embedding space. RelationNet \cite{yang2018learning} extended this line of work to use a separate non-linear comparison module instead of relying entirely on the embedding networks to make the data linearly separable  \cite{koch2015siamese,snell2017prototypical,vinyals2016matching}. This division of labour between a deep embedding and a deep relation module improved performance in practice \cite{yang2018learning}. Our approach builds on this line of work in general and RelationNet in particular. RelationNet relied on the embedding networks to produce a \emph{single} embedding for the relation module to compare. We argue that a general purpose comparison function should use any or all of the full feature hierarchy \cite{zeiler2014understandingCNN} to make matching decisions. For example matching based on colors, textures, or parts -- which may be represented at different layers in a embedding network. To this end we modularise the embedding networks, and pair every embedding module with its own relation module.

\keypoint{Use of Feature Hierarchies}
The general strategy of simultaneously exploiting multiple layers of a feature hierarchy has been exploited in conventional many-shot classification network \cite{huang2017densely,srivastava2015highwayNet}, instance recognition \cite{chang2018mlfn}, and semantic segmentation networks \cite{hariharan2015hypercolumn}. However, in the context of deep-metric learning, the conventional pipeline is to extract a complete feature \cite{ge2018deepMetric,hu2014deepMetric}. Importantly, in contrast to prior approaches single `short-cut' connection of deeper features to a classifier \cite{hariharan2015hypercolumn,chang2018mlfn}, we uniquely learn a hierarchy of relation modules: One non-linear comparison function for each block of the embedding modules. Our approach is also reminiscent of classic techniques such as spatial pyramids \cite{lazebnik2006pyramid} (since each module in the hierarchy operates at different spatial resolutions) and multi-kernel learning \cite{vedaldi2009mklObjDet} (since we learn multiple relation modules for each feature in the hierarchy). This can also be seen as the first multiple meta-learner approach for few shot learning problems.

\keypoint{Leaned Noise and Regularisation}
Many previous FSL models struggle with deeper backbones \cite{mishra2018simple, finn2017model}. For best performance, we would like to  exploit a state-of-the-art embedding module architecture (we use SENet \cite{hu2018senet}), and also benefit from the array of comparison modules mentioned above. To enable \modelnameshort{} to benefit from deep backbones without overfitting, we modify the embedding modules to output a feature distribution at each layer. Rather than generating deterministic features at a module output, we generate means and variances which are sampled in the forward pass, with back propagation relying on the reparamaterization trick. Unlike density networks \cite{bishop1994mdn} where such distributions are only generated at the output layer, or VAEs \cite{kingma2014variationalAutoEncoder} here they are generated only once by the generator, we generate  such stochastic features at \emph{each} embedding module's output. This can be seen as an end-to-end learnable data augmentation strategy in semantic feature rather than image space. It is also complementary to standard L2/weight decay and image space augmentation techniques.

\section{Methodology}
\subsection{Problem Definition} 
We consider a $C$-way $K$-shot classification problem for few shot learning. There are some labelled source tasks with sufficient data, denoted meta-train $\mathcal{D}_{\text{m-train}}$, and we ultimately want to solve a new set of target tasks denoted meta-test $\mathcal{D}_{\text{m-test}}$, for which the label space is disjoint.  Within meta-train and meta-test, we denote each task as being composed of a support set of training examples, and a query set of testing examples. The meta-test tasks are assumed to be few-shot, so $\mathcal{D}_{\text{m-test}}$ contains a support set with $C$ categories and $K$ examples each. We want to learn a model on meta-train that can generalize out of the box, without fine-tuning, to learning the new categories in meta-test.

\begin{figure}[h]
\centering
\includegraphics[width=0.9\columnwidth]{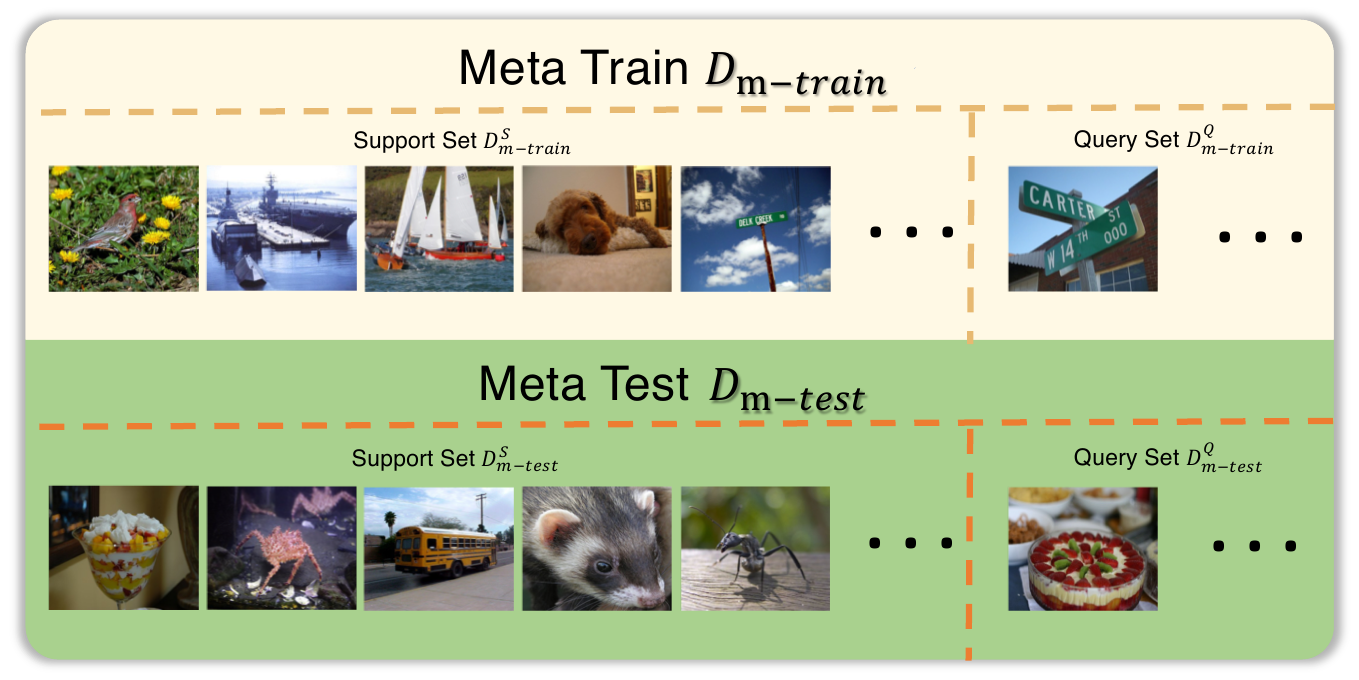}
\caption{\small Few-shot learning: Problem Setup.}
\label{fig:meta}
\end{figure}

\keypoint{Episodic Training} 
We adopt an episodic training paradigm for few-shot meta-learning. During meta-training, an episode is formed as follows: (i) Randomly select $C$ classes from $\mathcal{D}_{\text{m-train}}$, (ii) Sample $K$ images each class, which serve as \emph{support set} $ {\mathcal{D}}_{\text{m-train}}^{\text{S}} = \left\{ (x_{i}, y_{i})\right\}_{i=1}^m $, where $m=K*C$, (iii) For the same $C$ classes, sample $K'$ images each class serving as the \emph{query set} $ {\mathcal{D}}_{\text{m-train}}^{\text{Q}} = \left\{ (\tilde{x}_{j}, \tilde{y}_{j})\right\}_{j=1}^n$, where $n=K'*C$, $\mathcal{D}_{\text{m-train}}^{\text{S}} \cap \mathcal{D}_{\text{m-train}}^{\text{Q}} = \emptyset$. The support/query distinction mimics the  $\mathcal{D}_{\text{m-test}}$/ real-time testing. Our few-shot \modelnameshort{}  will be trained for instance comparison using episodes constructed in this manner. 

\subsection{Model}
\label{sec:model}
\begin{figure*}[t]
\centering
\includegraphics[width=0.99\textwidth]{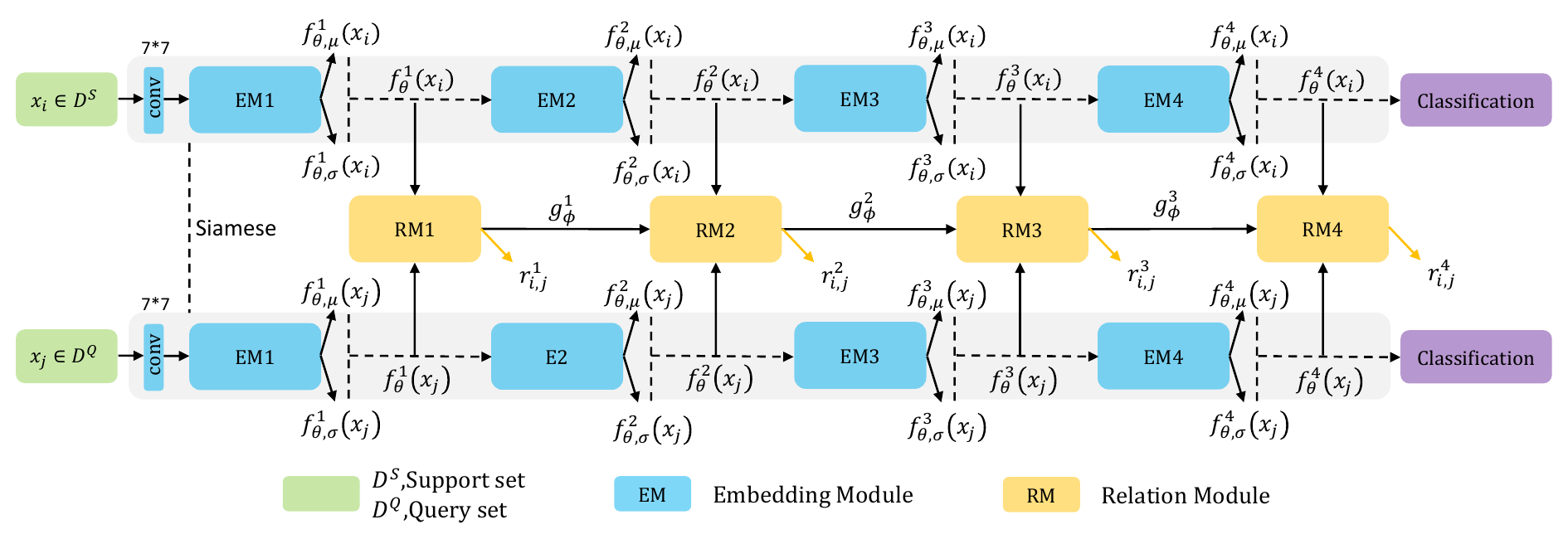}
\caption{\small RelationNet2's DCN architecture. There are 4 embedding modules $f_\theta$ for each embedding branch, and a set of 4 corresponding relation modules $g_\phi$. Support set and query set share the same embedding network. Each embedding module outputs a feature distribution $\mathcal{N}(f_{\theta,\mu}(x),f_{\theta,\sigma}(x))$, we then randomly sample a feature $f_{\theta}(x)$ as the input of corresponding relation module and next embedding module.\cut{ Embedding modules $f$ and relation modules $g$ are paramaterized by $\theta$ and $\psi$ respectively.}}
\label{fig:network}
\end{figure*}

\keypoint{Overview} RelationNet2's 
\modelnamefull is composed of two module types: \emph{embedding} and \emph{relation} modules $f_\theta$ and $g_\phi$, as shown in Fig.~\ref{fig:network}. The detailed architecture will be given in Section~\ref{sec:arch}. A pair of images $x_{i}$ and $x_{j}$ in the support and query set are fed to embedding modules respectively. Then the multi-level embedding modules output stochastic features to the corresponding multi-level relation modules, and learn the relation score and weights for different relation modules. Finally, the \modelnameshort{} learns weighted non-linear metric of few shot learning tasks.

\keypoint{Distribution Embedding Modules}
Conventionally, an embedding module (e.g., a ResNet or SENet block) outputs deterministic features. As a regularisation strategy, we treat each feature output as a random variable drawn from a parameterized Gaussian distribution, for which the embedding module outputs the mean and variance. This design is illustrated in Fig.~\ref{fig:network}. Each $v$th-level embedding module predicts a feature mean $f^v_{\theta,\mu}$ and a feature variance $f^v_{\theta,\sigma}$. To generate a module's output $f^v_\theta$, we use the reparameterization trick to draw one (or more) Gaussian random samples 
\begin{equation}
f_{\theta}^{v} = f^v_{\theta,{\mu}} + \varepsilon \odot f^v_{\theta,{\sigma}},
\label{noise}
\end{equation}
\noindent where $\varepsilon$ is a standard Gaussian  $\mathcal{N}(0,1)$ random samples, and $\odot$ denotes element-wise product. 

\keypoint{Metric Hierarchy}
The $v$th-level of embedding modules produce query and support image feature maps, which are concatenated as $[f_{\theta}^{v}(x_{i}), f_{\theta}^{v}(x_{j})]$, and then fed into the corresponding $v$th-level relation module for comparison.

For a pair $x_i$ and $x_j$ at level $v-1$, the relation module outputs a similarity feature map $g_{\phi}^{v-1}$. The $v$th-level relation module takes both the $v$th-level embedding output for query and support, and also the $(v-1)$th-level relation module similarity feature map as input:

\begin{equation}
\begin{aligned}
g_{\phi}^{v}= g([f_{\theta}^{v}(x_{i}),f_{\theta}^{v}(x_{j}),g_{\phi}^{v-1}]).
\end{aligned}
\label{relation_score}
\end{equation}

The first relation module is special as it does not have a predecessor to input, and we cannot use zero-padding because $0$ has a specific meaning in our context. Thus we set $g_{\phi}^{1}=g([f_{\theta}^{1}(x_{i}),f_{\theta}^{1}(x_{j})])$.

Simultaneously, after an average pooling and fully connected layer denoted $q(\cdot)$, each relation module also outputs a real-valued scalar representing similarity/relation score $r_{i,j}^v$ of two images estimated at the feature level $v$,
\begin{equation}\begin{aligned}
r_{ij}^v = q(g_{\phi}^{v}).
\end{aligned}\label{relation_score}\end{equation}
\keypoint{K-Shot} For $K$-shot with $K>1$, the embedding module outputs the average pooling of features, along the sample axis, of all samples from the same class to produce \emph{one} feature map. Thus, the number of outputs for the $v$-level relation module is $C$, regardless of the value of $K$. 

\keypoint{Objective Function} 
There are 2 steps to train the DCN network. We first train the embedding network, then fix the embedding network parameters and train the relation network (run the whole 
\modelnameshort{} consisting of embedding and relation modules, but only update the relation modules).

We first train the embedding network $\theta$ as a conventional multi-class classifier for the data in $\mathcal{D}_{m-train}$ using cross entropy loss $\ell^{CE}$. To leverage our distribution-embedding, we add a feature variance regularizer:

\begin{equation}
\theta \leftarrow \underset{\theta}{\operatorname{argmin}}~ \ell^{CE}(\theta)  - \lambda \frac{1}{m} \sum_{i=1}^m \sigma_i,
\label{celoss}
\end{equation}
where $\sigma_i$ is the predicted standard deviation of each instance and $m$ is their total number, and $\lambda$ is the hyperparameter to finetune the influence of the regularizer (here is 0.01). This ensures that feature distributions are learned, and we do not collapse to standard (zero-variance) vector embedding (our mean $\sigma$ is about 0.5). This pipeline can be seen as a learnable data augmentation strategy at each level of the feature hierarchy for relation modules. Learning with these augmented features improves generalization. After embedding training, the parameters $\theta$ of embedding modules are fixed.

We next train the column of relation modules $\phi$ on  $\mathcal{D}_{m-train}$ with an episodic strategy \cite{vinyals2016matching} using cross entropy loss $\ell^{CE}$ at each module (Fig.~\ref{fig:network}). To weight the $V$ relation modules, we assign a learnable attention weight $w^v_{c,j}$ to the calculated relation similarity score $r_{c,j}^v$ of each module. 

\begin{equation}
\phi \leftarrow  \underset{\phi}{\operatorname{argmin}}~  \sum_{c=1}^C \sum_{j=1}^n \sum_{v=1}^V \ell^{CE}({w_{c,j}^{v}{r_{c,j}^v}, \mathbf{1}(y_{c}=y_{j});\phi)},
\end{equation}
where $j=1\dots n$ refers to query samples and $c$ refers to a batch of $K$ support examples of class $y_c$ in a $C$-way-$K$-shot problem. $r_{c,j}$ are the relation scores between query image $j$ and the class $y_c$ support images.

Additionally, $w^v_{c,j}=\alpha^v(g^v_{c,j})$ is a sigmoid-activated fully connected layer that computes a scalar attention weight given relation feature map $g^v_{c,j}$, and the weights of $\alpha^v$ are included in $\phi$.

\keypoint{Testing Strategy} 
To evaluate our learned model on $C$-way-$K$-shot learning, we calculate the final relation score $r_{c,j}$ of one query image $x_j$ to the images of each support class $c$:

\begin{equation}
r_{c,j}=\cut{\sum_{i=1}^K}\sum_{v=1}^V{w}_{j}^{v}{r_{c,j}^v}
\label{eq:score}
\end{equation}
where $r^v_{c,j}$ is the relation score between image $j$ and the support images of class $c$ at module $v$.

Finally, the class with the highest relation score $r_{c}$ is the final predicted classification. We evaluate the approach by the resulting classification accuracy.

\subsection{Network Architecture}
\label{sec:arch}

RelationNet2's \modelnameshort{} architecture (Fig.~\ref{fig:network}) uses $4$ embedding modules, each paired with a relation module. We explain our method with SENet for concreteness, but it can be instantiated with any backbone.

\keypoint{Embedding Subnetwork} 
As shown in Fig.~\ref{fig:network}, first we use a $7\times 7$ convolution followed by a $3\times 3$ max-pooling, which is a common size reduction as \cite{hu2018senet}. Then, we have $4$ embedding modules each composed of a number of SENet blocks. Finally, an avg-pooling and a fully-connected layer are used to produce $C$ logit values, corresponding to $C$ classes in $\mathcal{D}_{\text{m-train}}$. More specifically, 4 embedding modules followed the 4 SENet basic blocks composition $[3,4,6,3]$, respectively.
In original SENet paper \cite{hu2018senet}, they use SE-ResNet-50, but here we use smaller backbones as SE-ResNet-34, where $(3+4+6+3)*2+2=34$. Otherwise, we follow the other setting in \cite{hu2018senet}, e.g., reduction ratio $r=16$ as suggested.

\keypoint{Distribution Embedding}
Conventually, an embedding module outputs deterministic features. As explained in  Section~\ref{sec:model}, each DCN embedding module's output is split into two parts: the mean feature $f_{\theta,{\mu}}$ sized $[b,c,h,w]$ ([batch\_size, channel, height, width]), and standard deviation (std) $f_{\theta,{\sigma}}$ sized $[b,1,h,w]$. 

We assume that every channel shares the same standard deviation (std). This means, in addition to the penultimate-to-output layer (now it is penultimate-to-mean layer), we have a new penultimate-to-std layer (with its own parameters). The motivation behind sharing stds across channels is to reduce the number of parameters in the newly introduced layer. We also control the amount of noise added by applying Sigmoid activation to constrain the std to the range $[0,1]$. We sample one feature vector per image in a single forward pass, but multiple samples are drawn considering the whole batch.

\keypoint{Relation Subnetwork} As illustrated in Fig.~\ref{fig:network}, the relation column consists of $4$ serial modules, each of which has $2$ SENet blocks, with a pooling and a fully-connected layer to produce the relation score. Thus the relation modules is designed as [2,2,2,2], where the SENet block architecture is the same as the one used in embedding module.

\section{Experiments}
We evaluate RelationNet2's \modelnameshort{} architecture on few-shot classification with \miniIN{} and \tierIN{} datasets. PyTorch code to reproduce results is available at \url{https://github.com/zhangxueting/DCN}. 

\keypoint{Baselines} 
We compare several state-of-the-art baselines for few-shot learning including
Matching Nets \cite{vinyals2016matching},
Meta Nets \cite{munkhdalai2017meta},
Meta LSTM \cite{ravi2017optimization}, 
MAML \cite{finn2017model},
Baseline++ \cite{chen2019closerfewshot},
Prototypical Nets \cite{snell2017prototypical}, 
Graph Neural Nets \cite{garcia2017few}, 
Meta-SSL \cite{ren2018meta}, 
Relation Net \cite{yang2018learning}, 
Meta-SGD \cite{li2017meta}, 
TPN \cite{liu2018transductive},
CAVIA \cite{zintgraf2018cavia},
DynamicFSL \cite{gidaris2018dynamic},
SNAIL \cite{mishra2018simple},
AdaResNet \cite{munkhdalai2018rapid},
TADAM \cite{oreshkin2018tadam},
MTL \cite{sun2019meta},
TapNet \cite{yoon2019tapnet}, 
MetaOpt Net \cite{lee2019meta}, 
PPA \cite{qiao2017few}, 
LEO \cite{rusu2019leo}.

\keypoint{Data Augmentation}
We follow the standard data augmentation \cite{szegedy2015going,hu2018senet, he2016deep, chen2019closerfewshot} with random-size cropping and random horizontal flipping
Input images are normalized through mean channel subtraction.

\cut{\keypoint{Pre-train and train}
The embedding branch is pre-trained by the training set and the parameters then are fixed. The validation set is used to estimate the number of early stop episodes for the relation training. Finally, both train and validation data (as per common practice \cite{qiao2017few}) are used to train the relation modules in \modelnameshort{}.}

\subsection{\textit{mini}Imagenet}
\keypoint{Dataset}
\miniIN{} has 60,000 images in consist of 100 \imagenet{} classes, each with 600 images  \cite{vinyals2016matching}. Following the split in \cite{ravi2017optimization}, the dataset is divided into a 64-class training set, 16-class validation set and a 20-class testing set.

\keypoint{Settings}
We evaluate both \textit{5-way-1-shot} and \textit{5-way-5-shot}, where each episode contains 5 query images for each sampled class. There are 5*5+1*5=30 images per training episode/mini-batch for 5-way-1-shot experiments, and 5*5+5*5=50 images for 5-way-5-shot experiments. When it comes to 5-shot, we calculate the class-wise average feature across the support set. Thus we get 5*5*5*1=125 feature pairs as input for the relation module.
For embedding and relation module training, optimization uses SGD with momentum 0.9. The initial learning rate is 0.1, decreased by a factor of 5 every 60 epochs, and the training epoch is 200. All models are trained from scratch, using the robust RELU weight initialization \cite{he2015delving}. We follow \cite{chen2019closerfewshot} in using 224$\times$224 pixels crops for evaluation on ResNet and SENet, and \cite{ravi2017optimization} in using 84$\times$84 images for the smaller Conv-4 backbone.

\keypoint{Results}
Following the setting of \cite{snell2017prototypical}, when evaluating testing performance, we batch 15 query images per class in a testing episode and the accuracy is calculated by averaging over 600 randomly generated testing tasks (for both 1-shot and 5-shot scenarios). 
\cut{Firstly, our idea of learned Gaussian noise for better regularisation and multiple metrics can be applied to other metric learning methods (e.g. ProtoNet \cite{snell2017prototypical}), We extended ProtoNet to use four metrics instead of one, and also adding the learned Gaussian Noise for regularisation. Also, for fair comparison, we control different embedding backbones. The results of 5-way-1-shot in Tab~\ref{tab:first} show that while learned noise and multiple metrics also improve the performance of ProtoNet, \modelnameshort{} (RelationNet with 4 metrics and learned noise) is still the best.}
In Tab.~\ref{tab:mini}, \modelnameshort{} achieves excellent performance with different embedding backbones. Specifically, the accuracy of 5-way \miniIN{} with SENet is 63.19\% and 76.58\% for 1-shot and 5-shot respectively. We note that MetaOptNet \cite{lee2019meta} uses significantly more advanced regularizers than standard among the competitors (which corresponds to about $2\%$ performance according to \cite{lee2019meta}), also requires an order of magnitude higher dimensionality of embeddings [64,160,320,640] than the other competitors [64,96,128,256]. Overall DCN's 1-shot recognition performance is state-of-the-art among methods that do not require optimisation at meta-test time (unlike, e.g., MAML \cite{finn2017model} and MetaOptNet \cite{lee2019meta}). It is noteworthy that achieving good performance with deeper backbones is not trivially automatic as Dynamic FSL, for example fails to improve from Conv-4 to ResNet embedding. DCN's learned noise regularizer helps it to exploit a powerful SENet backbone without overfitting. Direct comparison among models is complicated by the diversity of embedding networks used in different studies, so we show the results of DCN with each commonly used backbone in Tab.\ref{tab:mini}, e.g. Conv-4 and ResNet-12. We can see that DCN performs favorably across a range of architectures.

\setlength{\tabcolsep}{4.8pt}
\begin{table}[t]
\centering
\footnotesize
\begin{tabular}{@{} llcc @{}}
\toprule
\multirow{2}{*}{\bf Model}  &\multirow{2}{*}{\bf Embedding} &\multicolumn{2}{c}{\multirow{2}{*}{\bf \textit{mini}Imagenet 5-way Acc.}}\\
& \multicolumn{2}{c}{}  \\
&& 1-shot & 5-shot \\
\midrule 

\textbf{\textsc{Matching} \textsc{Nets}} \cite{vinyals2016matching} &Conv-4 & 43.56 $\pm$ 0.84\% &55.31 $\pm$ 0.73\%  \\ 
\textbf{\textsc{Meta} \textsc{LSTM}} \cite{ravi2017optimization} &Conv-4  &43.44 $\pm$ 0.77\% & 60.60 $\pm$ 0.71\% \\ 
\textbf{\textsc{MAML}}$^O$ \cite{finn2017model} & Conv-4 &  48.70 $\pm$ 1.84\% & 63.11 $\pm$ 0.92\% \\ 
\textbf{\textsc{Baseline++}} \cite{chen2019closerfewshot} & Conv-4 &  48.24 $\pm$ 0.75\% & 66.43 $\pm$ 0.63\% \\ 
\textbf{\textsc{Meta} \textsc{Nets}} \cite{munkhdalai2017meta} & Conv-5 &49.21 $\pm$ 0.96\% & - \\
\textbf{\textsc{ProtoNet}} \cite{snell2017prototypical} & Conv-4 &49.42 $ \pm $ 0.78\% &68.20 $\pm$ 0.66\%  \\ 
\textbf{\textsc{GNN}} \cite{garcia2017few} & Conv-4 & 50.33 $\pm$ 0.36\% & 66.41 $\pm$ 0.63\% \\ 
\textbf{\textsc{Meta SSL}}\cite{ren2018meta}&  Conv-4 &50.41 $\pm$ 0.31\% & 64.39 $\pm$ 0.24\% \\ 
\textbf{\textsc{Relation} \textsc{Net}} \cite{yang2018learning}& Conv-4 & 50.44 $\pm$ 0.82\% &65.32 $\pm$ 0.70\% \\ 
\textbf{\textsc{Meta SGD}}$^O$ \cite{li2017meta}  & Conv-4 & 50.47 $\pm$ 1.87\% & 64.03 $\pm$ 0.94\% \\ 
\textbf{\textsc{TPN}} \cite{liu2018transductive} & Conv-4 & 52.78 $\pm$ 0.27\% & 66.59 $\pm$ 0.28\% \\ 
\textbf{\textsc{CAVIA}} \cite{zintgraf2018cavia} & Conv-4 & 51.82 $\pm$ 0.65\% & 65.85 $\pm$ 0.55\%\\
\textbf{\textsc{Dynamic FSL}}$^\dagger$ \cite{gidaris2018dynamic} & Conv-4 & \textbf{56.20 $\pm$ 0.86\%} & \textbf{72.81 $\pm$ 0.62\%}\\
\textbf{\textsc{RelationNet2 (\modelnameshort{})}} & Conv-4 & {53.48 $\pm$ 0.78\%} & {67.63 $\pm$ 0.59\%} \\ 

\midrule
\textbf{\textsc{Baseline++}} \cite{chen2019closerfewshot} & ResNet-18 &  51.87 $\pm$ 0.77\% & 75.68 $\pm$ 0.63\% \\ 
\textbf{\textsc{RelationNet}} \cite{chen2019closerfewshot} & ResNet-18 &  52.48 $\pm$ 0.86\% & 69.83 $\pm$ 0.68\% \\
\textbf{\textsc{ProtoNet}} \cite{chen2019closerfewshot} & ResNet-18 &  54.16 $\pm$ 0.82\% & 73.68 $\pm$ 0.65\% \\ 
\textbf{\textsc{SNAIL}} \cite{santoro2017simple} & ResNet-12 &55.71 $\pm$ 0.99\% & 68.88 $\pm$ 0.92\% \\ 
\textbf{\textsc{Dynamic FSL}} \cite{gidaris2018dynamic} & ResNet-12& 55.45 $\pm$ 0.89\% & 70.13 $\pm$ 0.68\% \\ 
\textbf{\textsc{adaResNet}} \cite{munkhdalai2018rapid} & ResNet-12 & 57.10 $\pm$ 0.70\% & 70.04 $\pm$ 0.63\% \\
\textbf{\textsc{TADAM}} \cite{oreshkin2018tadam} & ResNet-12 & 58.50 $\pm$ 0.30\% & {76.70 $\pm$ 0.30\%} \\
\textbf{\textsc{MTL}} \cite{sun2019meta} & ResNet-12$^*$ & 61.20 $\pm$ 1.80\% & 75.50 $\pm$ 0.80\% \\
\textbf{\textsc{Tap Net}} \cite{yoon2019tapnet} & ResNet-12 & 61.65 $\pm$ 0.15\% & 76.36 $\pm$ 0.10\% \\
\textbf{\textsc{MetaOptNet}}$^{O}$ \cite{lee2019meta} & ResNet-12$^*$ & \textbf{64.09 $\pm$ 0.62\%} & \textbf{80.00 $\pm$ 0.45\%} \\ 
\textbf{\textsc{RelationNet2 (\modelnameshort{})}} & ResNet-12 & \textbf{63.92 $\pm$ 0.98\%} & \textbf{77.15 $\pm$ 0.59\%} \\ 
\midrule

\textbf{\textsc{PPA}} \cite{qiao2017few} & WRN-28-10 & 59.60 $\pm$ 0.41\% & 73.74 $\pm$ 0.19\% \\ 
\textbf{\textsc{LEO}}$^{O}$ \cite{rusu2019leo} & WRN-28-10 & 61.78 $\pm$ 0.05\% & 77.59 $\pm$ 0.12\% \\ 

\midrule
\textbf{\textsc{MAML}} & SENet & 55.99 $\pm$ 0.99\%  & -\\
\textbf{\textsc{RelationNet}} & SENet & 57.39 $\pm$ 0.86\% & -\ \\ 
\textbf{\textsc{ProtoNet}} & SENet & 51.60 $\pm$ 0.85\% & -\ \\
\textbf{\textsc{RelationNet2 (\modelnameshort{})}} &SENet & \textbf{63.19 $\pm$ 0.87\%} & \textbf{76.58 $\pm$ 0.66\%} \\ 

\bottomrule
\end{tabular}
\caption{\small \small
Few-shot classification results on \miniIN{}. All accuracies are averaged over 600 test episodes and are reported with 95\% confidence intervals. Best-performing method is bold, along with any others whose confidence intervals overlap. From top to bottom: Simple conv block embeddings to other deep embeddings (ResNet, WRN, SENet). `-': not reported. $^\dagger$: use two-step optimization with added attention. \cut{$^\#$: Trained with union of train and validation classes.} $^O$: requires gradient-based optimisation at meta-test time. $^*$: Use a wider ResNet than standard and higher dimensional embedding. 
}
\label{tab:mini}
\end{table}

\keypoint{Cross-way Testing Results}
Standard procedure in few-shot evaluation is to train models for the desired number of categories to discriminate at testing time. However, unlike alternatives such as MAML \cite{finn2017model}, our method is not required to match label cardinality between training and testing. We therefore evaluate 5-way trained model on 20-way testing in Tab.~\ref{tab:ablation2}. It shows that our model outperforms the alternatives clearly despite \modelnameshort{} being trained for 5-way, and the others specifically for 20-way, indicating another important aspect of \modelnameshort{}'s flexibility and general applicability. 

\setlength{\tabcolsep}{4.8pt}
\begin{table}[t]
\centering
\footnotesize
\begin{tabular}{@{} llcc @{}}
\toprule
\multirow{2}{*}{\bf Model}  & \multirow{2}{*}{\bf Embedding}  & \multicolumn{2}{c}{\multirow{2}{*}{\bf \textit{mini}Imagenet 20-way Acc.}}\\
& \multicolumn{2}{c}{}  \\
& & 1-shot & 5-shot \\
\midrule 
\textbf{\textsc{Matching Nets}} \cite{li2017meta} & Conv-4 & 17.31 $\pm$ 0.22\% &  22.69 $\pm$ 0.86\%  \\ 
\textbf{\textsc{Meta LSTM}}$^O$ \cite{li2017meta}& Conv-4 & 16.70 $\pm$ 0.23\% &  26.06 $\pm$ 0.25\%   \\
\textbf{\textsc{MAML}}$^O$ \cite{li2017meta}& Conv-4 & 16.49 $\pm$ 0.58\%  &  19.29 $\pm$ 0.29\%   \\
\textbf{\textsc{Meta SGD}}$^O$ \cite{li2017meta}& Conv-4 & 17.56 $\pm$ 0.64\%  &  28.92 $\pm$ 0.35\% \\ 
\midrule  
\textbf{\textsc{RelationNet2 (\modelnameshort{})}} & Conv-4 &27.56 $\pm$ 0.24\% & 39.56 $\pm$ 0.81\%\\
\textbf{\textsc{RelationNet2 (\modelnameshort{})}}  & ResNet-12 &31.65 $\pm$ 0.34\% & 50.25 $\pm$ 0.46\%\\
\textbf{\textsc{RelationNet2 (\modelnameshort{})}} & SENet &\textbf{32.90 $\pm$ 0.39\%} & \textbf{51.37 $\pm$ 0.39\%} \\ 

\bottomrule
\end{tabular}%
\caption{\small \small
20-way classification accuracy on \miniIN{}. \modelnameshort{} is trained on 5-way with different embeddings and transferred to 20-way. Meta LSTM, MAML, and Meta SGD results are from \cite{li2017meta}. 
}
\label{tab:ablation2}
\end{table}

\subsection{\textit{tiered}Imagenet}
\keypoint{Dataset}
\tierIN{} is a larger few-shot recognition benchmark containing 608 classes (779,165 images), in which training/validation/testing categories are organized so as to ensure a larger semantic gap than those in \miniIN{}, thus providing a more rigorous test of generalization. This is achieved by dividing according to 34-higher-level nodes in the ImageNet hierarchy \cite{ren2018meta}, grouped into 20 for training (351 classes), 6 for validation (97 classes) and 8 for testing (160 classes), respectively.

\keypoint{Settings}
Similar to the setting of \miniIN{}, we use 5 query images per training episode. Due to the larger data size, we train embedding modules with a larger batch size 512, initial learning rate 0.3 and 100 training epochs. Other settings remain the same as \miniIN{}.

\keypoint{Results}
Following the former experiments, we batch 15 query images per class in each testing episode and the accuracy is calculated by averaging over 600 randomly generated testing tasks. From Tab.~\ref{tab:tiered}, \modelnameshort{} achieves the state-of-the-art performance on the 5-way-1-shot and 5-shot tasks with comfortable margins. Again, this is state-of-the-art performance for methods that do not require optimisation at meta-testing. We note also that Meta-SSL \cite{ren2018meta} and TPN \cite{liu2018transductive} are semi-supervised methods that use more information than ours, and have additional requirements such as access to the test set for transduction.

\setlength{\tabcolsep}{4.8pt}
\begin{table}[t]
\centering
\footnotesize
\begin{tabular}{@{} llcc @{}}
\toprule
\multirow{2}{*}{\bf Model}  &\multirow{2}{*}{\bf Embedding} &\multicolumn{2}{c}{\multirow{2}{*}{\bf \textit{tiered}Imagenet 5-way Acc.}}\\
& \multicolumn{2}{c}{}  \\
&& 1-shot & 5-shot \\
\midrule 

\textbf{\textsc{Reptile}}  \cite{liu2018transductive}& Conv-4 &48.97\%  &66.47\% \\
\textbf{\textsc{MAML}} \cite{liu2018transductive} & Conv-4&  51.67\% & 70.30\% \\ 
\textbf{\textsc{Meta SSL}}$^\dagger$ \cite{ren2018meta}&Conv-4& 52.39 $\pm$ 0.44\% & 70.25 $\pm$ 0.31\% \\
\textbf{\textsc{Proto Net}} \cite{liu2018transductive} &Conv-4&53.31\%  &72.69\% \\ 
\textbf{\textsc{Relation} \textsc{Net}} \cite{liu2018transductive} &Conv-4& 54.48\% &71.31\% \\ 
\textbf{\textsc{TPN}}$^\dagger$ \cite{liu2018transductive}&Conv-4& 59.91\% & 73.30\% \\

\textbf{\textsc{Tap Net}} \cite{yoon2019tapnet} & ResNet-12& 63.08 $\pm$ 0.15\% & 80.26 $\pm$ 0.12\% \\

\textbf{\textsc{MetaOptNet}}$^O$ \cite{lee2019meta} & ResNet-12$^*$ & 65.81 $\pm$ 0.74\% & \textbf{ 81.75 $\pm$ 0.53\% }\\
\midrule
\textbf{\textsc{RelationNet2 (\modelnameshort{})}} & Conv-4 & 60.58 $\pm$ 0.72\% & 72.42 $\pm$ 0.69 \% \\
\textbf{\textsc{RelationNet2 (\modelnameshort{})}} & ResNet-12 & 68.58 $\pm$ 0.63\%  & \textbf{80.65 $\pm$ 0.91\%}\\

\textbf{\textsc{RelationNet2 (\modelnameshort{})}} & SENet & \textbf{68.83 $\pm$ 0.94\%} & 79.62 $\pm$ 0.77\% \\ 

\bottomrule
\end{tabular}%
\caption{\small \small
Few-shot classification results on \tierIN{}. All accuracies are averaged over 600 test episodes and reported with 95\% confidence intervals. For each task, the best-performing method is bold. $^\dagger$: Make use of additional unlabeled data for semi-supervised learning or transductive inference.  $^O$: requires gradient-based optimisation at meta-test time. $^*$: Uses a wider ResNet than standard size and higher dimensional embedding. 
}
\label{tab:tiered}
\end{table}

\setlength{\tabcolsep}{4.8pt}
\begin{table}[t]
\centering
\footnotesize
\begin{tabular}{@{} lc @{}}
\toprule
{\bf Model}  & {\bf \miniIN{} 5-way-1-shot Acc.}\\
\midrule 
\textbf{\textsc{\modelnameshort{}}} Full model  & {63.19} $\pm$ 0.87\%  \\ 
\textbf{\textsc{\modelnameshort{}}}-No module weight & {62.88} $\pm$ 0.83\%  \\ 
\textbf{\textsc{\modelnameshort{}}}-No deep sup.   &58.02 $\pm$ 0.80\% \\ 
\midrule
\textbf{\textsc{\modelnameshort{}}}-$r_{1}$ & 52.25 $\pm$ 0.80\%  \\ 
\textbf{\textsc{\modelnameshort{}}}-$r_{2}$  & 58.07 $\pm$ 0.80\%  \\ 
\textbf{\textsc{\modelnameshort{}}}-$r_{3}$  & 60.69 $\pm$ 0.81\%  \\ 
\textbf{\textsc{\modelnameshort{}}}-$r_{4}$  & 58.31 $\pm$ 0.79\%  \\
\bottomrule
\end{tabular}%
\caption{\small \small
Ablation study using 5-way-1-shot classification on \miniIN{} evaluating the impact of regularization techniques and multiple relation modules.
}
\label{tab:ablation1}
\end{table}


\begin{figure*}[t]
    \centering
    \includegraphics[width=0.24\textwidth]{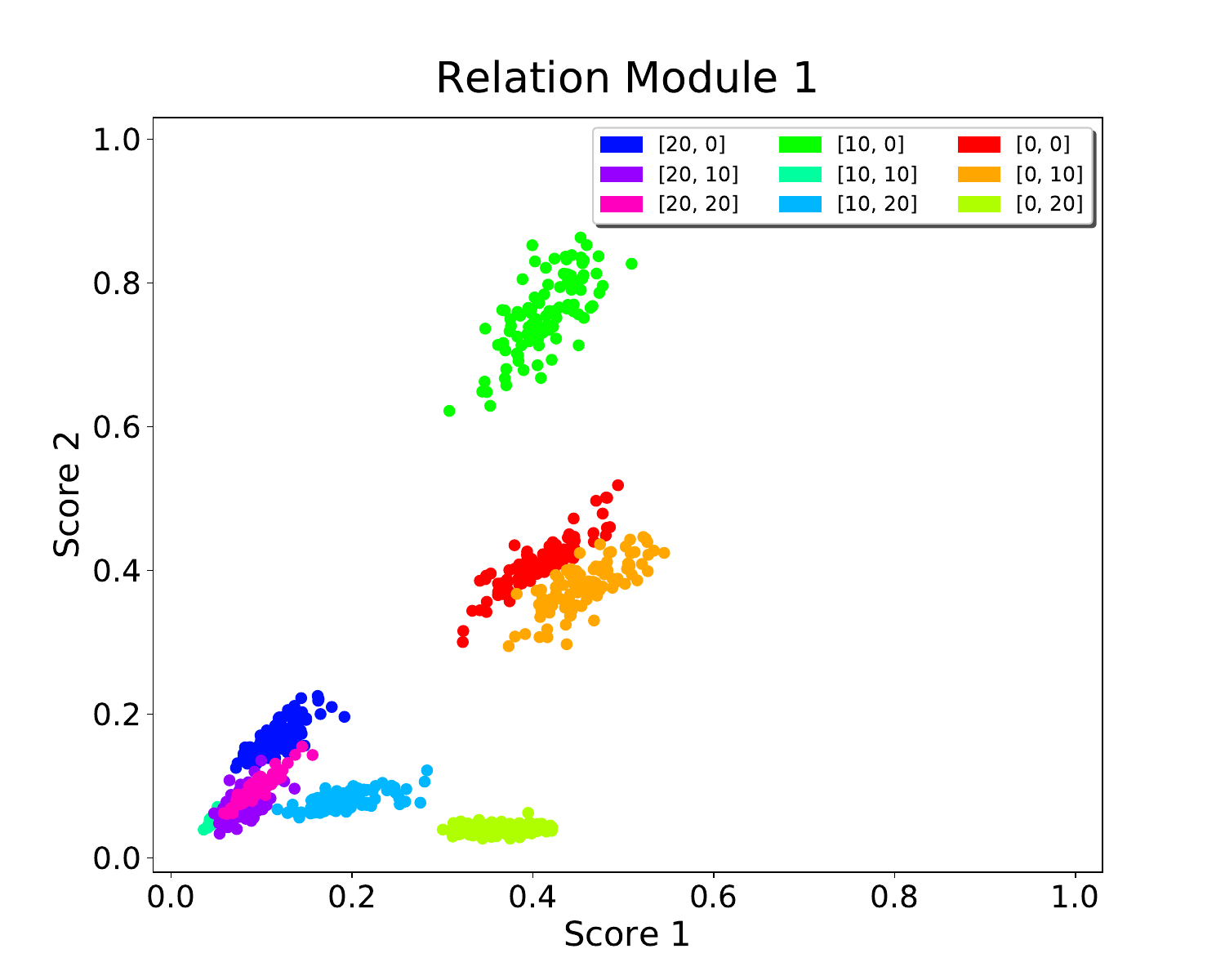}
    \includegraphics[width=0.24\textwidth]{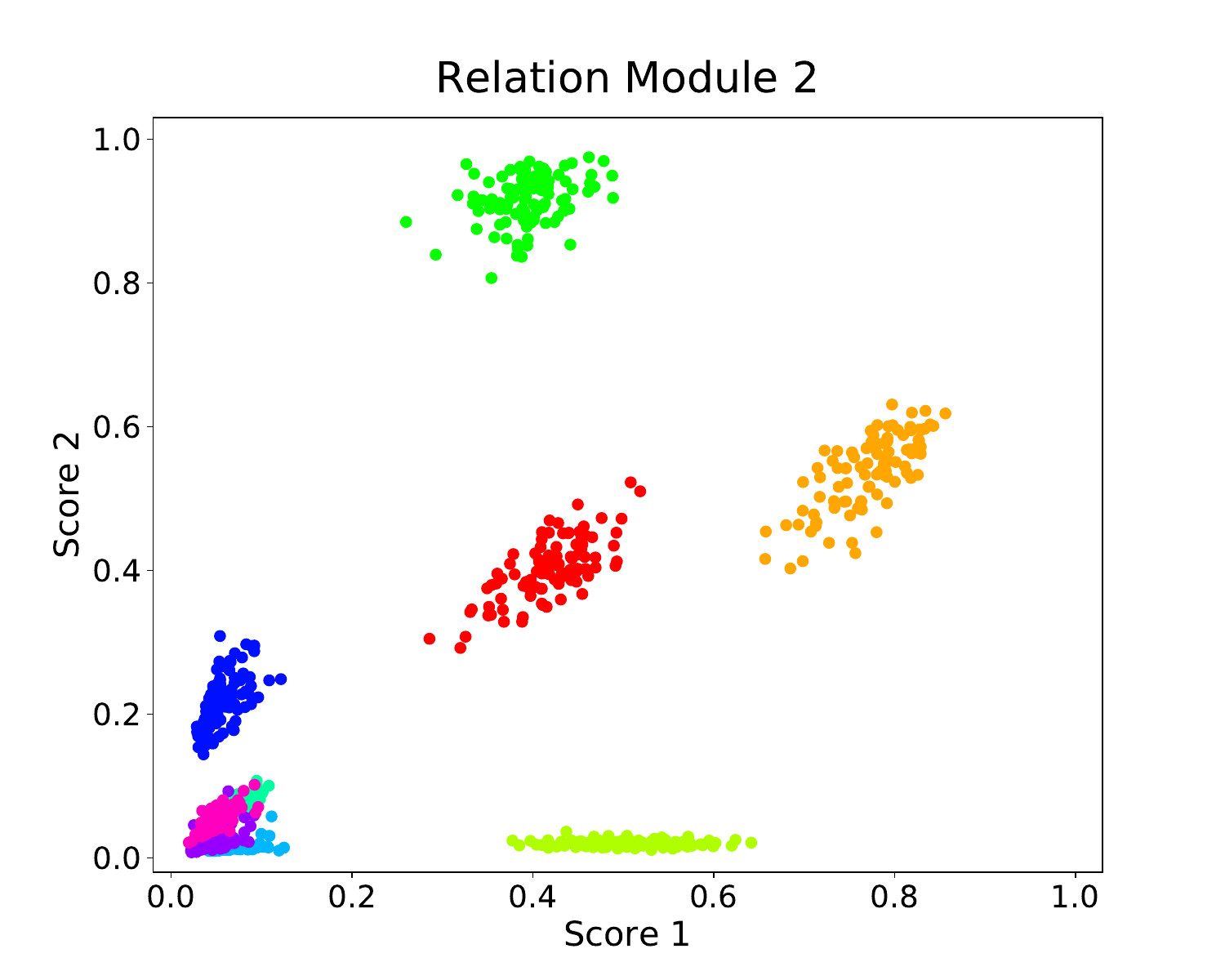}
    \includegraphics[width=0.24\textwidth]{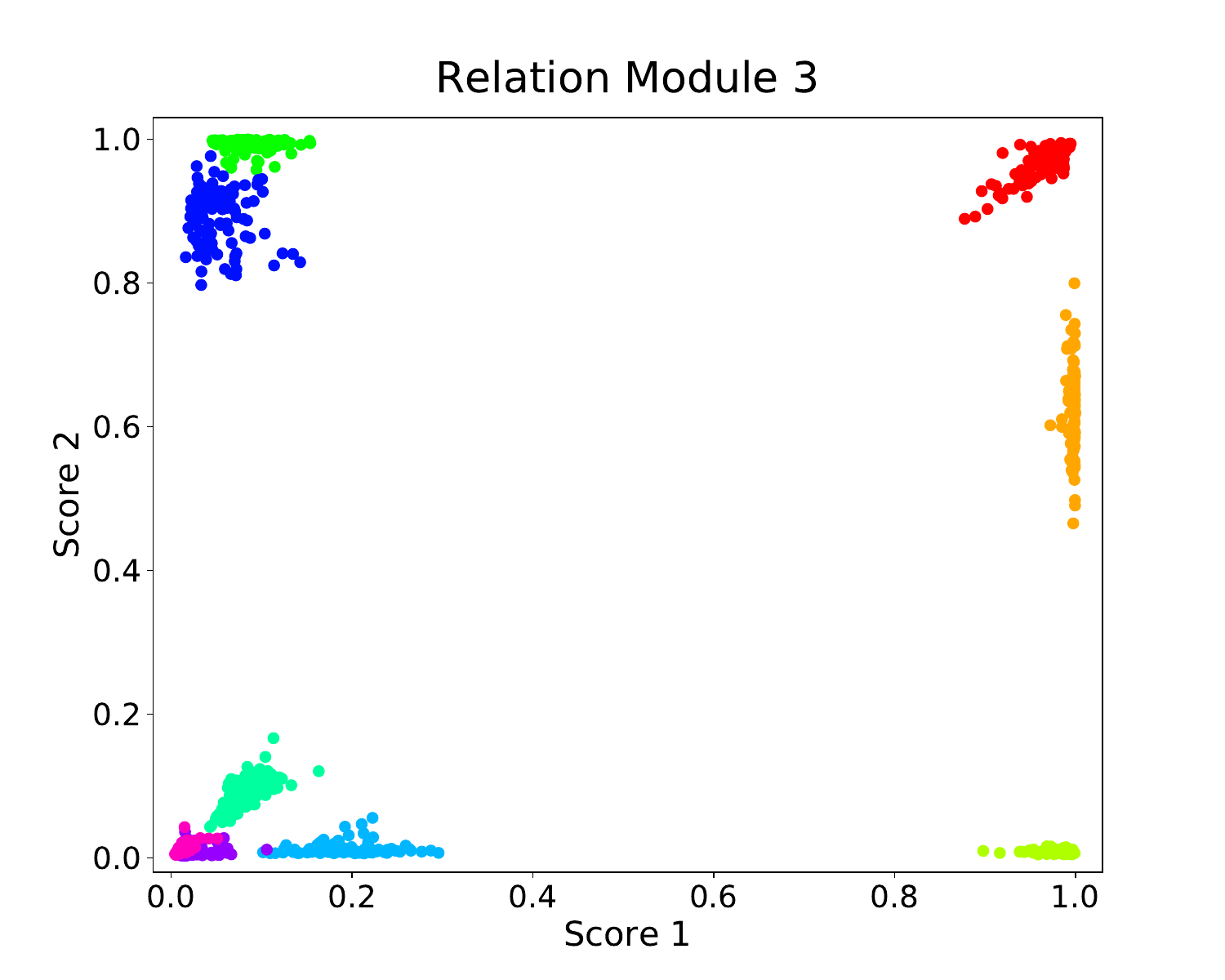}
    \includegraphics[width=0.24\textwidth]{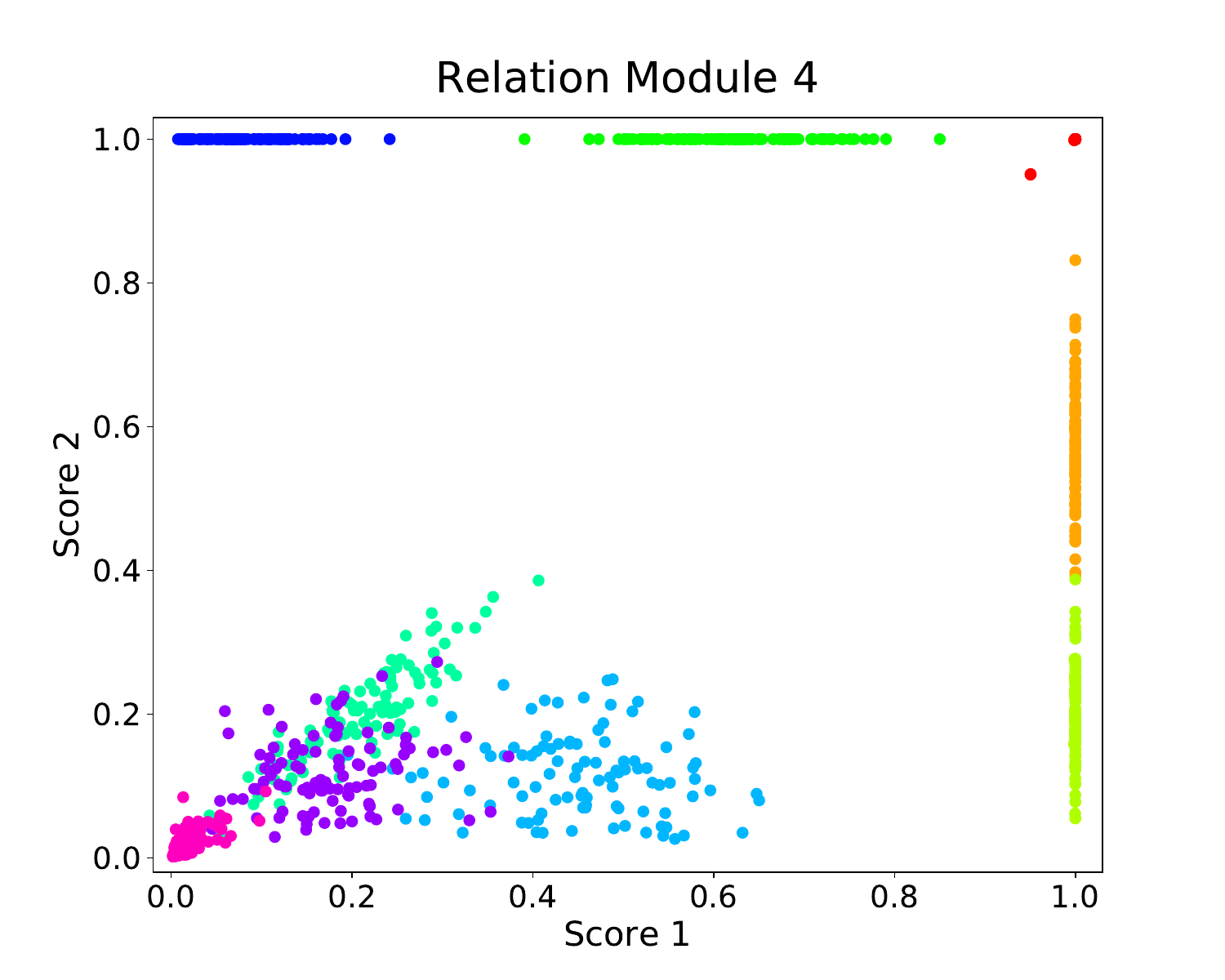}
    \vspace{-0.1cm}
    \caption{\small Illustration of query-support score distribution and the link to \imagenet{} hierarchy. Colors indicate query images of a $(query,support1,support2)$ class triple matching the specified \imagenet{} distance relationship $[D(q,s1),D(q,s2)]$.  \cut{{\color{blue} $\hrectangleblack$} $[20,10]$, {\color{green} $\hrectangleblack$} $[10,0]$, {\color{red} $\hrectangleblack$} $[0,0]$}}
    \label{fig:my_label}
\end{figure*}

\begin{figure}[tb]
    \centering
    \includegraphics[width=0.9\columnwidth]{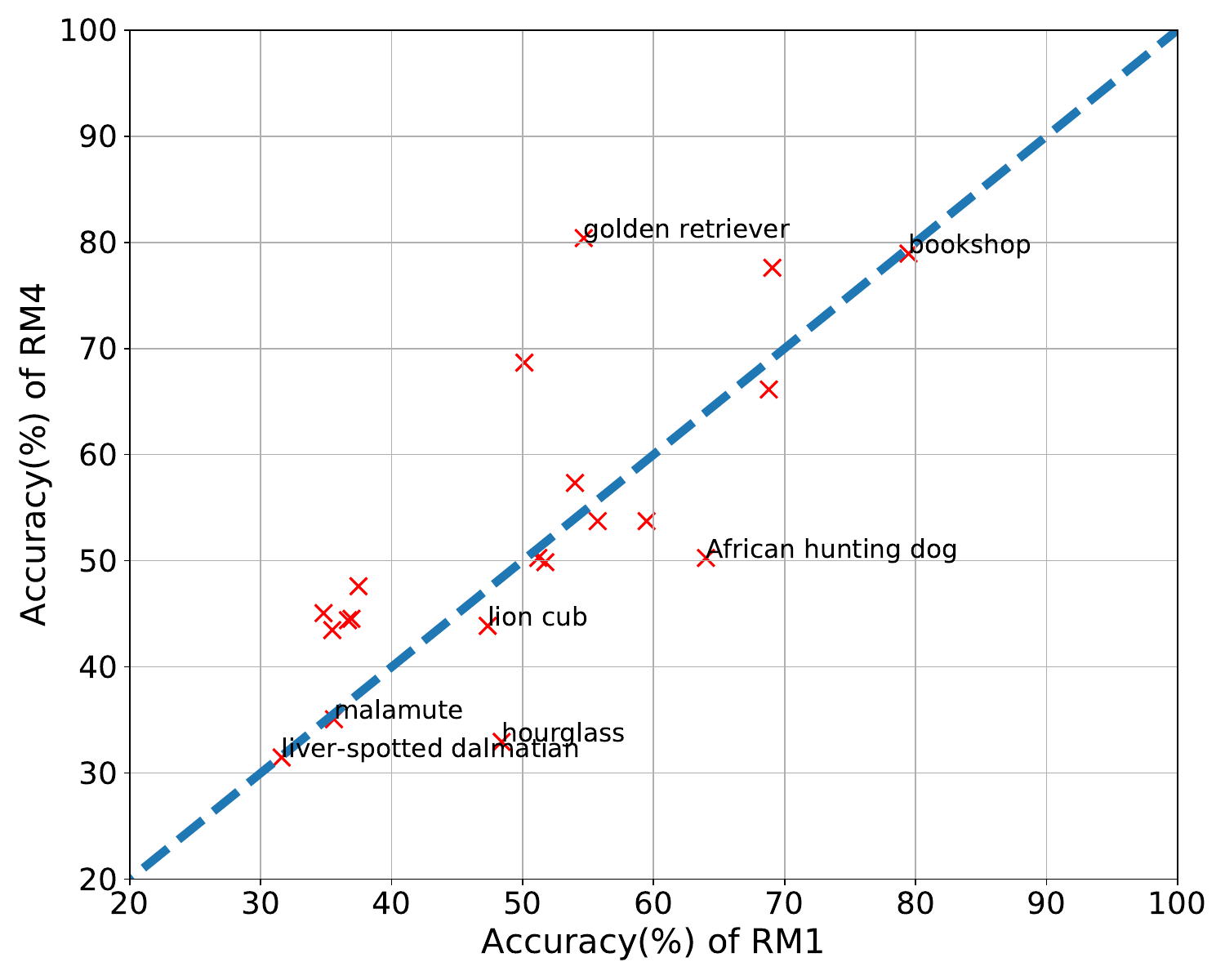}
    \vspace{-0.2cm}
    \caption{\small Category-wise accuracy of RM1 vs RM4. Different relation modules are better at detecting different categories. }
    \label{fig:accScatter}
\end{figure}

\setlength{\tabcolsep}{4.8pt}
\begin{table}[tb]
\centering
\footnotesize
\begin{tabular}{@{} cccccc @{}}
\hline
{\bf Module} & {\bf RM1} & {\bf RM2} & {\bf RM3} & {\bf RM4} \\
\midrule
\textbf{RM1} & - & - & - & - \\
\textbf{RM2} & 0.75 & - & - & - \\
\textbf{RM3} & 0.55 & 0.73 & - & - \\
\textbf{RM4} & 0.34 & 0.45 & 0.61 & - \\
\bottomrule
\end{tabular}
\caption{\small Spearman rank-order correlation coefficient between different relation modules: Modules make diverse predictions.}
\label{tab:score-corelation}
\end{table}

\subsection{Further Analysis}\label{sec:further}

\subsubsection{Application to Other Metric Learners} Our main insight is the value of feature comparison at multiple abstraction levels in metric learning, as well as that of learned noise regularizers for deep networks in the few-shot regime. We now confirm that these ideas can be applied to other base metric learners. Tab~\ref{tab:first} shows the 5-way-1-shot \miniIN{} results for both RelationNet \cite{yang2018learning} and ProtoNet \cite{snell2017prototypical} base learners controlling for these features. We can see that both architectures benefit from deep comparisons and regularizers. However the benefit is greater for RelationNet, which we attribute to the learnable non-linear  relation modules. These can learn a different comparison function at each abstraction level, but are also more complex so benefit more from the additional regularisation.

\setlength{\tabcolsep}{3.8pt}
\begin{table}[t]
\centering
\footnotesize
\begin{tabular}{@{}lccc @{}} 
\toprule
{\bf Model} & {\bf Noise Reg.?} & {\bf Deep Comparisons?} & {\bf Acc.}\\
\midrule 
\textbf{\textsc{ProtoNet}}\cite{snell2017prototypical} \cut{\cite{snell2017prototypical}} & \text{\sffamily X} & \text{\sffamily X} - 1 module & 51.04 $\pm$ 0.77\% \\ 
\textbf{\textsc{ProtoNet}} & \checkmark & \text{\sffamily X} - 1 module & 51.60 $\pm$ 0.85\% \\ 
\textbf{\textsc{ProtoNet}} & \text{\sffamily X} & \checkmark - 4 modules & 53.62 $\pm$ 0.82\% \\
\textbf{\textsc{ProtoNet}} & \checkmark & \checkmark - 4 modules & 54.78 $\pm$ 0.88\% \\
 
\midrule
\textbf{\textsc{RelationNet}}\cite{yang2018learning} & \text{\sffamily X} & \text{\sffamily X} - 1 module  & 52.48 $\pm$ 0.86\% \\ 
\textbf{\textsc{RelationNet}} & \checkmark &  \text{\sffamily X} - 1 module & 57.39 $\pm$ 0.86\%  \\ 
\textbf{\textsc{RelationNet2 (\modelnameshort{})}} & \text{\sffamily X} &\checkmark - 4 modules& 60.57 $\pm$ 0.86\% \\
\textbf{\textsc{RelationNet2 (\modelnameshort{})}} & \checkmark & \checkmark - 4 modules & 63.19 $\pm$ 0.87\% \\
\bottomrule
\end{tabular}
\caption{\small \small
Multiple deep comparisons and distribution embedding of features benefit both RelationNet (learnable relation modules) and ProtoNet (fixed linear modules) few-shot architectures. Accuracy is calculated on 5-way-1-shot classification of \textit{mini}Imagenet.
}
\label{tab:first}
\end{table}

\subsubsection{Ablation Study}
We further investigate the detailed design parameters of our method with a series of ablation studies reported in Tab.~\ref{tab:ablation1}. The conclusions are as follows:

\cut{\textbf{Parameterized Gaussian Noise Regularization:} Comparing \modelnameshort{} and \modelnameshort{}-No Noise, we can see that this brings over 2\% improvement.} \cut{\textbf{Retraining:} The impact of re-training on the combined training and validation set is visible by comparing the entries with \modelnameshort{}-No Retrain. Retraining provides a similar 2\% margin, and this is complementary to the noise.} \textbf{Deep Supervision:} The \modelnameshort{}-No Deep Sup. result shows that deep supervision is important to gain full benefit from a column of relation modules.  \textbf{Module Weighting:} Learning attention weights per module helps somewhat compared to manually tuned module weights. More importantly it eliminates the need for hand-tuning model weights. \textbf{Multiple Non-linear Metrics:} Tab~\ref{tab:ablation1} also shows the testing accuracy with each \modelnameshort{} relation module output score $r_v$ in isolation (\modelnameshort{}-$r_v$). Each module performs competitively, but their combination clearly leads to the best overall performance, supporting our argument that multiple levels of the feature hierarchy should be used to make general purpose matching decisions. Multiple meta learner design is a creative contribution of our work. 

\subsubsection{Architecture}
Our \modelnameshort{} benefits from deeper embedding architectures (Tab.~\ref{tab:mini}). It improves when going from simple convolutional blocks (used by early studies \cite{finn2017model,snell2017prototypical,yang2018learning}), to ResNet \cite{he2016deep} and SENet \cite{hu2018senet}. For fair comparison, when fixing a common ResNet-12, \modelnameshort{} outperforms the others that do not require meta-test optimization. Moreover, when fixing a common SENet, competitors RelationNet/ProtoNet/MAML are improved, but still surpassed by \modelnameshort{}.  

\cut{Our idea of multiple metrics can be applied to other metric learning methods. We extended Prototypical Nets to use four (linear) metrics instead of one, analogous to our use of four comparison modules in \modelnameshort{}. The results in Tab~\ref{tab:mini} (bottom) show that while multiple metrics also improve the performance of Prototypical Nets, \modelnameshort{} is still much better.}  


\subsubsection{Relation Module Analysis}
A key contribution in \modelnameshort{} is to perform metric learning at multiple abstraction levels simultaneously via a series of paired relation and embedding modules. Relation modules are analyzed to provide insight into the complementarity.

\keypoint{Score-Distance Correlation} We first check how the relation module (RM) scores relate to distances in the ImageNet hierarchy. Using \miniIN{} data, we search for $(support1,support2,query)$ category tuples where the distance $D(query,support1)$ and $D(query,support2)$ match a certain number of links, and then plot instances from these tuples query categories against the relative relation module scores $RM(q,s1)$, $RM(q,s2)$. Fig.~\ref{fig:my_label} presents scatter plots for the four relation modules where points are images and colors indicate category tuples with specified distance from the two support classes. We can see that: (1) The scores generally match \imagenet{} distances: The most/least similar categories (red/magenta) are usually closer to the top right/bottom left of the plot; while query categories closer to one support class are in the opposite corners (blue/yellow-green). (2) Generally higher numbered relation modules are more discriminative, separating classes with larger differences in relation score. 

\keypoint{Score Correlation} We next investigated if relation module predictions are diverse or redundant. We analyzed the correlation in their predictions by randomly picking 10,000 image pairs from \miniIN{} and computing the Spearman rank-order correlation coefficient \cite{Spearman1904Proof} between each pair of relation module's scores. The results in Tab.~\ref{tab:score-corelation}, show that: (1) Many correlations are relatively low (down to 0.34), indicating that they are making diverse, non-redundant predictions; and (2) Adjacent RMs have higher correlation than non-adjacent RMs, indicating that prediction diversity is related to RM position in the feature hierarchy.

\keypoint{Prediction Success by Module} We know that RM predictions do not necessarily agree. But to find out if they are complementary, we made a scatter plot of the per-class accuracy of RM-1 vs RM-4 in Fig.~\ref{fig:accScatter}. We can see that many categories lie on the diagonal, indicating that RM-1 and-4 get them right equally often. However there are some categories \emph{below} the diagonal, indicating that RM-1 gets them right more often than RM-4. Examples include both stereotyped and fine-grained categories such as `hourglass' and `African hunting dog'. These below diagonal elements confirm the value of using deeper features in metric learning.

\section{Conclusion}
We proposed RelationNet2, a general purpose matching framework for few-shot learning. It implements a \modelname{} architecture that performs effective few-shot learning via learning multiple non-linear comparisons  corresponding to multiple levels of feature extraction, while resisting overfitting through stochastic regularisation. The resulting method achieves state-of-the-art results on \miniIN{} and the more ambitious \tierIN{}, while retaining architectural simplicity, and fast training and testing processes.

\keypoint{Acknowledgements} This work was supported by EPSRC grant EP/R026173/1. 

{\small
\bibliographystyle{IEEEtran}
\bibliography{IEEEabrv, egbib}
}

\clearpage

\end{document}